\documentclass[10pt,twocolumn,letterpaper]{article}

\usepackage{iccv}
\usepackage{times}
\usepackage{epsfig}
\usepackage{graphicx}
\usepackage{amsmath}
\usepackage{amssymb}
\usepackage{caption} 
\usepackage[pagebackref=true,breaklinks=true,letterpaper=true,colorlinks,bookmarks=false]{hyperref}

\iccvfinalcopy 

\def\iccvPaperID{5307} 

\ificcvfinal

\begin{document}

\title{CLIP3Dstyler: Language Guided 3D Arbitrary Neural Style Transfer}
\author{
${Ming Gao}^{*1}$, ${YanWu Xu}^{*2}$, ${Yang Zhao}^{3}$, ${Tingbo Hou}^{3}$, \\ ${Chenkai Zhao}^{1}$, ${Mingming Gong}^{4}$
\\
    {\tt\small
    $^{1}{University of Pittsburgh}$,
    $^{2}{Boston University}$,
    $^{3}{Google}$,
    $^{4}{University of Melbourne}$
    }
}

\begin{figure}[t!]
\twocolumn[{
\maketitle
    \includegraphics[width=\textwidth]{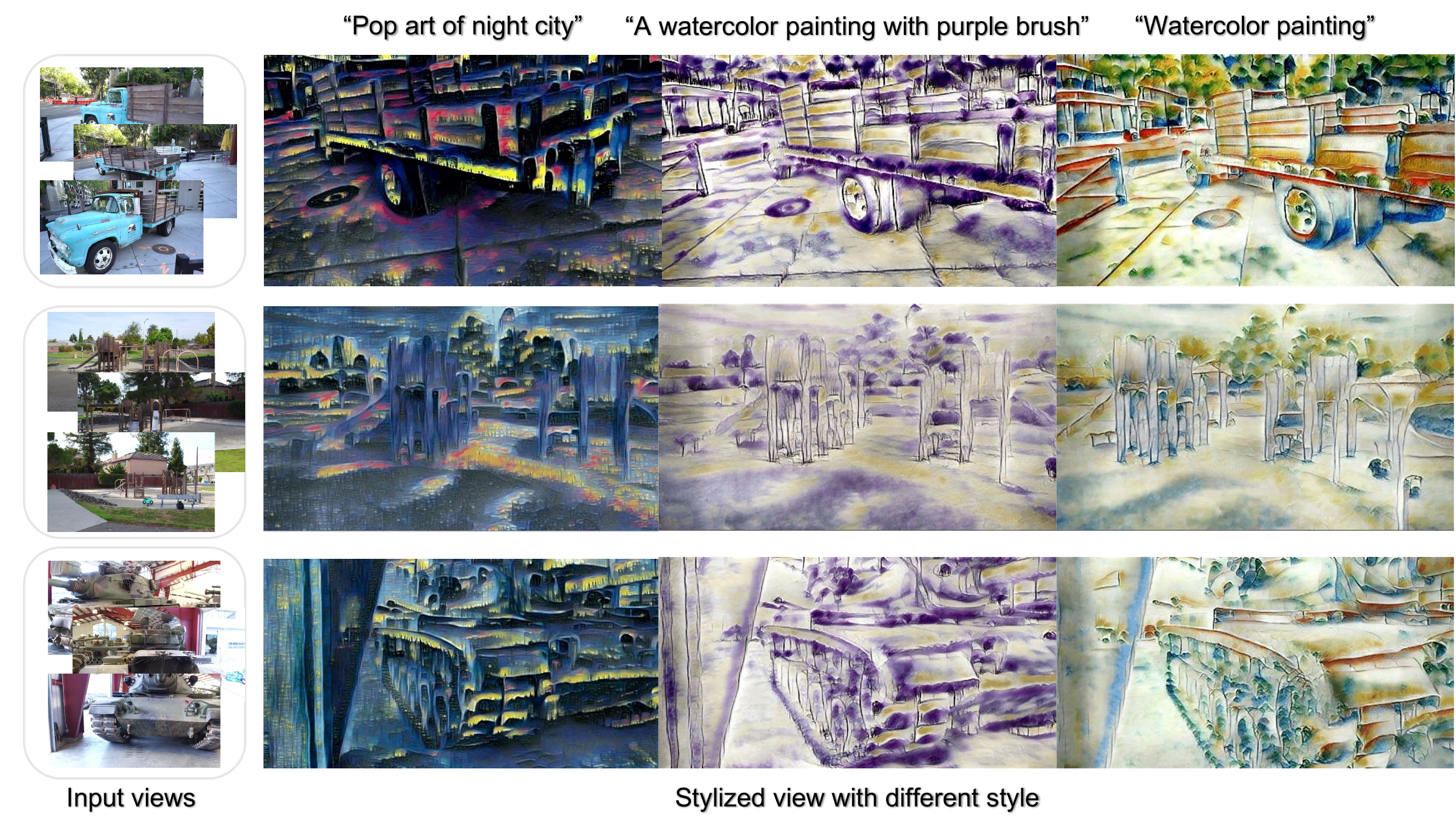}
    \caption{Given a set of image and text style information, with the 3D geometry information, our method achieves better style transfer results on various text conditions and keeps consistency across arbitrary novel views.  }
    \label{top}
}]
\end{figure}

\begin{abstract}
    In this paper, we propose a novel language-guided 3D arbitrary neural style transfer method (CLIP3Dstyler). We aim at stylizing any 3D scene with an arbitrary style from a text description, and synthesizing the novel stylized view, which is more flexible than the image-conditioned style transfer. Compared with the previous 2D method CLIPStyler, we are able to stylize a 3D scene and generalize to novel scenes without re-train our model. A straightforward solution is to combine previous image-conditioned 3D style transfer and text-conditioned 2D style transfer \bigskip methods. However, such a solution cannot achieve our goal due to two main challenges. First, there is no multi-modal model matching point clouds and language at different feature scales (\eg low-level, high-level). Second, we observe a style mixing issue when we stylize the content with different style conditions from text prompts. To address the first issue, we propose a 3D stylization framework to match the point cloud features with text features in local and global views. For the second issue, we propose an improved directional divergence loss to make arbitrary text styles more distinguishable as a complement to our framework. We conduct extensive experiments to show the effectiveness of our model on text-guided 3D scene style transfer.
\end{abstract}


\begin{figure*}[t!]
\centering
    \includegraphics[width=0.95\textwidth]{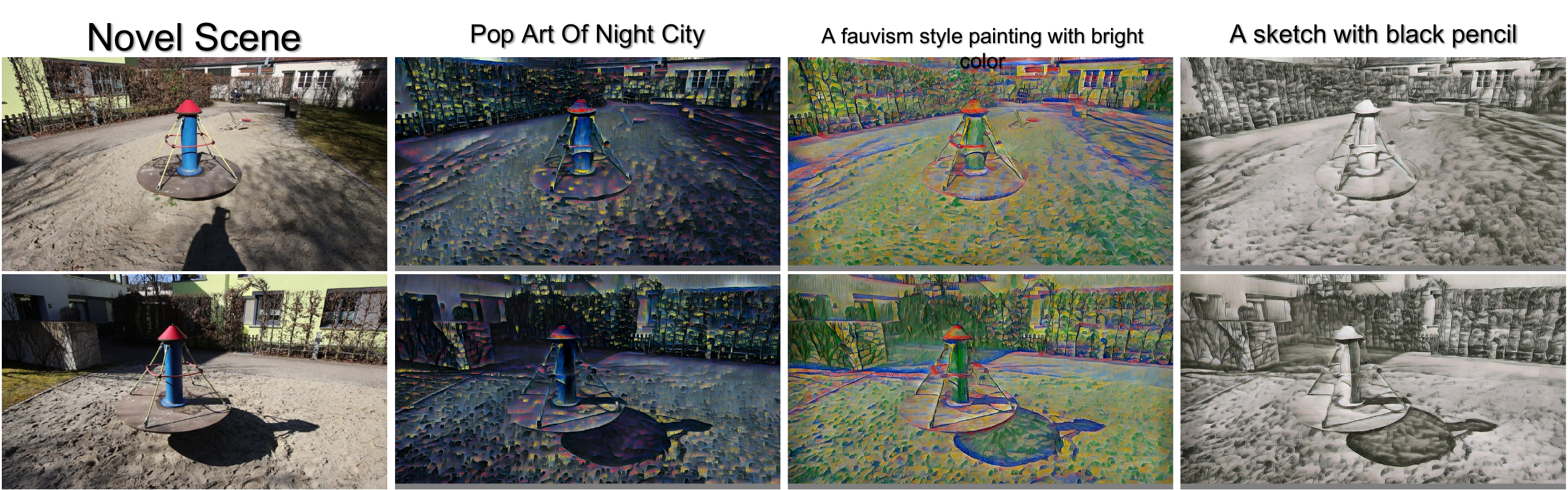}
    \caption{In this figure, we show that our model can be generalized to the hold-out test scenes without retraining our model on it.}
    \label{genelized_example}
    \vspace{-6mm}
\end{figure*}

\section{Introduction}
\label{sec:intro}
Vision-Language models \cite{kwon2021clipstyler,DBLP:journals/corr/abs-2103-00020,Patashnik_2021_ICCV,gal2021stylegannada,vinker2022clipasso} have shown superior advantages over most of the current tasks, 
such as semantic segmentation, object detection and action recognition. However, the 3D scenes stylised with the guidance of Vision-Language driven is rarely explored.

In this paper, we propose a 3D stylization framework that stylizes the 3D scenes via a given text description of the style, which can be applied to stylize novel scenes without further training. The proposed method will have multiple potential applications in the rising VR, AR, MetaVerse, etc., with more flexible user-defined features.

\begin{figure}[h!]
    \vspace{-2mm}
    \includegraphics[width=0.8\linewidth]{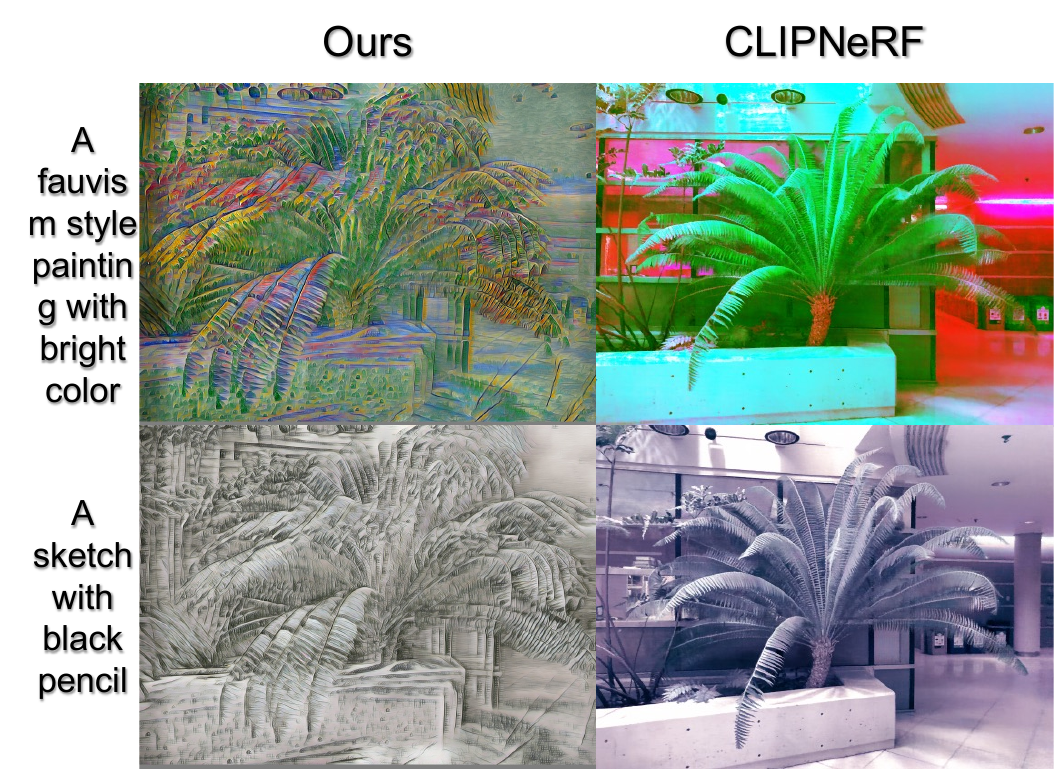}
    \caption{We also compare with the CLIPNeRF\cite{wang2021clip}; Note that we do not train the model on the nerf llff dataset; we only infer the model on these datasets and styles, but the CLIPNeRF needs to optimize for each scene and style.}
    \label{CLIPNeRF}
    \vspace{-8mm}
\end{figure}

The related work would be the arbitrary 3D scene stylization via a given style image \cite{gu2022stylenerf,huang_2021_3d_scene_stylization,mu20223d}, and the text-driven 2D image stylization \cite{kwon2021clipstyler,gal2021stylegannada,Patashnik_2021_ICCV}. The current 3D stylization work is built upon the point cloud 3D representations  \cite{mu20223d,huang_2021_3d_scene_stylization} or the recently popular Neural Radiance Field (NeRF) \cite{gu2022stylenerf}. Even though NeRF has several advantages over point clouds, such as easy training with only multiple views inputs and smoother interpolation between different views, the NeRF-based stylization models can only be applied to a single scene\cite{gu2022stylenerf}, which is inapplicable for stylizing multiple scenes or generalization on novel scenes, shown in Figure~\ref{genelized_example}. In the CLIPNeRF\cite{wang2021clip}, although the proposed 3D stylization method with text style condition shows a stable 3D consistency before and after stylization, it leads to a barely obvious style effect with the given text style condition (Figure~\ref{CLIPNeRF}), which is still under-explored for NeRF-based method. Thus, in this paper, we build our 3D CLIPStyler upon the point cloud representation. One of the key components of arbitrary style image stylization is to match the content and the style of the stylized images with the input content and the style images. To match the 3D feature and the given 2D style feature, \cite{mu20223d,huang_2021_3d_scene_stylization} project the 3D point descriptor to the 2D plane and transfer it to a 2D image feature matching problem with a pre-trained image encoder. For the text-driven 2D image stylization CLIPstyler \cite{kwon2021clipstyler} utilizes the image encoder and the text encoder of CLIP \cite{DBLP:journals/corr/abs-2103-00020} to match the image and text features in the same metric space.

In the under-explored text-driven 3D point cloud stylization task, we need to match the 3D point cloud descriptor with the text features. However, no such multi-modal model matches the feature of the point cloud and text caption. Generally, the natural solution is to bridge the ideas in \cite{huang_2021_3d_scene_stylization}, and \cite{kwon2021clipstyler} to project the 3D point cloud back to the 2D image space and match the text-image feature using CLIP. However, this straightforward solution faces two major challenges. First, the previous works refer to a style image to transfer the content images, where the content and the style are of the same modality. Thus, the multi-scale features extracted from different layers of a pre-trained VGG are all in the same feature space for the content, style, and stylized images, which is crucial for balancing faithful global style and local content details. However, in the pre-trained CLIP network, there is no such concept of a multi-scale feature for image-text matching; there is only a deep feature from the text encoder. In general, the deep style feature should transfer the deep content feature, but the point cloud descriptors are shallow layer's feature, which will cause blurred content and unfaithful style effect for stylizing novel views. Second, we observe that directly adopting the style matching loss of CLIPstyler \cite{kwon2021clipstyler} for multiple text styles transfer would lead to a mixing of style effects as shown in Figure~\ref{mixing}, which is undesirable for general arbitrary stylization. And this style mixing effect is the key factor preventing the model from learning different text styles. 

To address the above issues, we propose a more general framework for language-guided 3D point cloud stylization (CLIP3Dstyler). Our proposed CLIP3Dstyler enables the model to be trained with multiple scenes with arbitrary text styles and achieve content consistency and faithful style effect without mixing. To achieve content consistency and faithful style effect, we propose complimenting the local point cloud descriptor with a feature from the global view of the entire 3D scene to match the global text style feature. 
\begin{center}
    \centering
    \captionsetup{type=figure}
    \includegraphics[width=\linewidth]{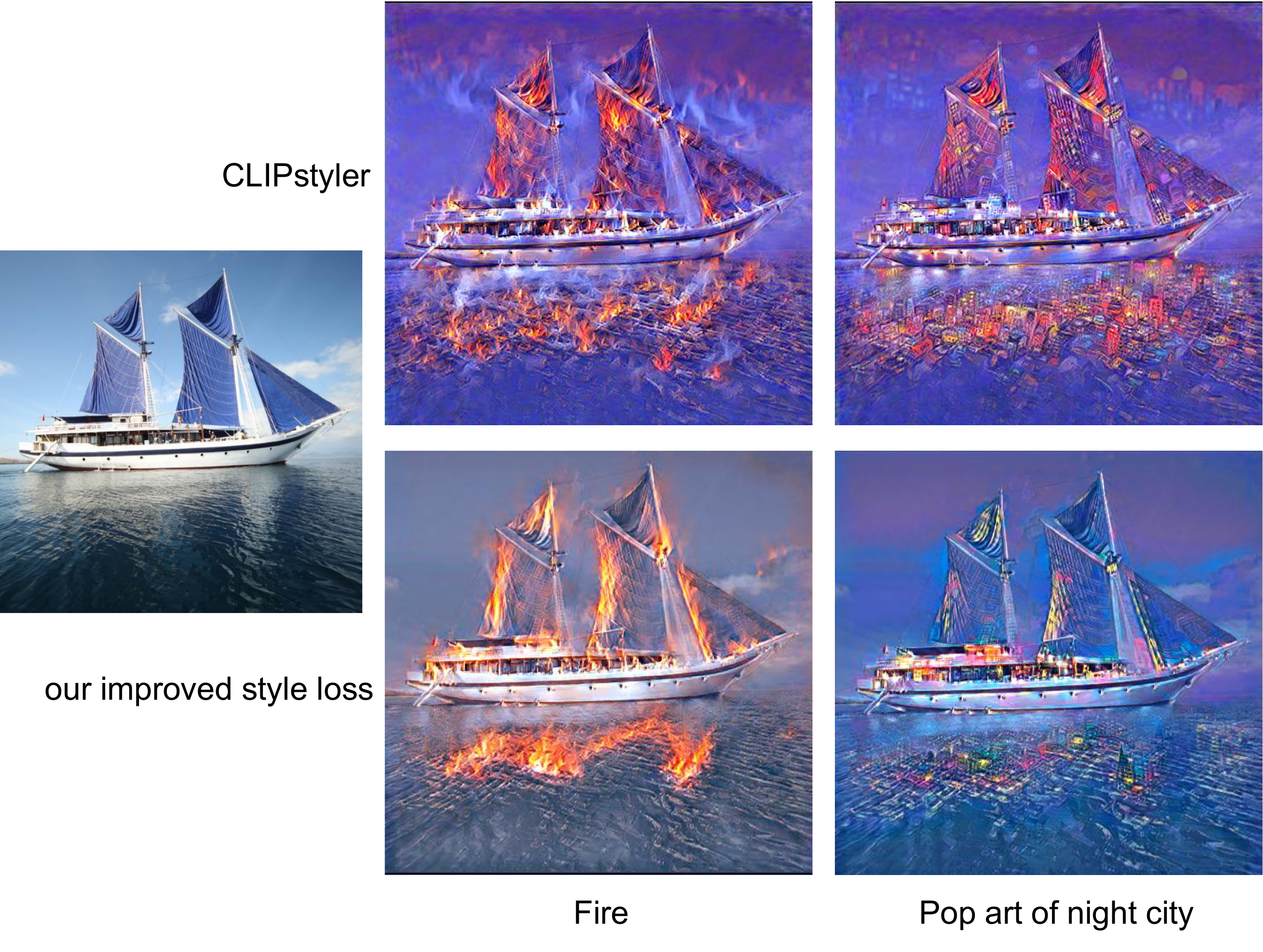}
    \captionof{figure}{Our proposed style divergence loss prevents the model from mixing styles}
    \label{mixing}
\end{center}%

\begin{figure*}[h!]
\centering
    \includegraphics[width=0.95\textwidth]{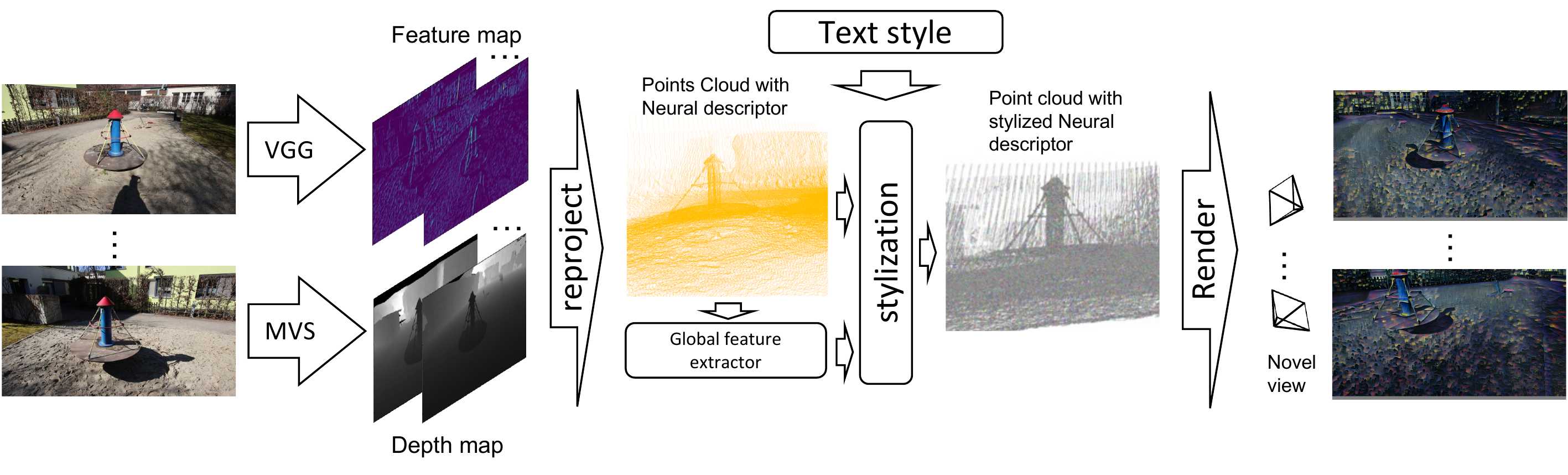}
    \caption{\textbf{Method Overview.} Our method starts from the point cloud reconstruction from a set of images and generate feature for each point. Point-wise and global feature representation stylized by given a text style. Integrating the per-points and global transferred features are projected to novel view and decoded to the RGB images.  }
    \label{architecture}
    \vspace{-6mm}
\end{figure*}
Furthermore, we designed a light module to extract the global feature, and the additional overhead can be ignored compared to our full model, which is efficient yet effective. To fix the style mixing problem when we have multiple styles, we propose an improved directional divergence loss for segregating the style effect from one to another, enhancing stylization significantly. Our model can also generalize to stylize novel scenes without retraining, as shown in Figure~\ref{genelized_example}.
Text styles. For more details, please refer to Section~\ref{method}
Our contribution is summarized into three points: 1) we present a new framework to address the language-guided 3D style transfer learning task; 2) We rethink the text style as a global information and generate associated global features from the point cloud to achieve better performance with a higher CLIP score; 3) We introduce a new directional divergence loss to solve the style mixing problem.

\section{Related Work}
\textbf{Language-driven model.} Recently, OpenAI introduced the CLIP\cite{DBLP:journals/corr/abs-2103-00020} pre-trained model to bridge the text and image modality. By contrastive Learning from 40 million image and text pairs, CLIP shows giant potential ability about the language conditional model and brings many exciting interaction research between the text and image. For example, StyleCLIP\cite{Patashnik_2021_ICCV} performs a different level of attribute manipulation from the text information of StyleGAN\cite{DBLP:journals/corr/abs-1812-04948}. DALL·E 2\cite{ramesh2022} integrates the CLIP and diffusion model to generate a high-resolution and vivid image. CLIPstyler\cite{kwon2021clipstyler} performs text-driven style transfer learning by the minimize the difference of directional distance of text and image feature extracted by CLIP encoder. Our task of text 3D stylization want to extend 2D style transfer to the 3D space, as it requires the synthesis of novel view, and keeps consistency between the different views.

\textbf{Novel view synthesis.} Multi-images or single images novel view synthesis can achieve by projection\cite{Meshry_2019_CVPR,Аliev2020}, and volume-based differentiate rendering. For volume rendering\cite{nerv2021,mildenhall2020nerf,tancik2022blocknerf,liu2020neural}, each pixel of a view image emits a ray of light. Each pixel gets the value in the picture by integrating the color and opacity value along the ray. The neural network is used as an implicit representation of 3D space. This neural network must be forward thousands of times in render process, which is the most severe bottleneck of render speed. Inspired by the traditional render pipeline, SynSin\cite{wiles2020synsin} proposed a projection based differentiate render pipeline by soft projecting the points onto a plane with a z-buffer to generate the feature map at the associate viewpoint. After the projection operation, it is attached with a decoder to create the render result. It is very efficient and capable of generalizing to arbitrary scene method. Hence, our model based on the projection differentiate rendering.

\textbf{3D stylization.} 3D content stylization\cite{Cao_2020_WACV,gu2022stylenerf,huang_2021_3d_scene_stylization,mu20223d} has attracted growing interest in recent years. StyleScene\cite{huang_2021_3d_scene_stylization} constructs the point cloud from a set of images and performs linear style transformation on each point associate feature. 3D Photo Stylization\cite{mu20223d} generates the point cloud by depth estimation from the single images and performs the GCN to extract the geometrical feature of the 3D scene. By Adaptive Attention Normalization(AdaAttN)\cite{liu2021adaattn}, the styles and contents are matched and combined by attention mechanism. With the post-back projection operation, novel views are synthesized. CLIPNeRF\cite{wang2021clip} use the neural radiance field as 3d representation, and text prompt as style information to perform stylization, but it can only change the color of the scene. In contrast, our approach achieves a very significant style migration effect.

\textbf{Deep learning for point cloud processing.} The deep neural network that gets the point cloud as input has been widely researched in classification\cite{qi2016pointnet,qi2017pointnetASplus,dgcnn,Wu_2019_CVPR}, semantic segmentation\cite{Zhao_2021_ICCV,1807.00652,rosu2020latticenet}, and object detection\cite{DBLP:journals/corr/abs-1812-05784,s18103337}. The voxel-based method\cite{DBLP:journals/corr/abs-1711-06396,DBLP:journals/corr/abs-1812-05784,s18103337} rasterizes the 3D space and gets the feature map for post-operation, but with the resolution increasing, memory consumption goes up exponentially. Point-Based\cite{qi2016pointnet,qi2017pointnetASplus,Zhao_2021_ICCV} methods directly take the point cloud into the network without the complex pre-processing technique and are more memory friendly. Our approach uses the Point-based method to extract the features from the point cloud to handle millions of points in a scene.
\section{Method}\label{method}
Given a set of 3D scenes of point clouds $ {\{P_n\}^N_{n=1}} $ and text style description $\{S_m\}_{m=1}^M$, our goal is to stylize the given 3D point clouds with arbitrary text style and synthesize novel stylization views of 3D scenes. To briefly introduce our training framework. We decompose the proposed method into three components as shown in Figure~\ref{architecture}; the first component generates a point cloud of the features via projecting image features into the corresponding 3D coordinates (depth map). The second component comprises a light global point cloud feature extractor and one off-the-shelf prompt feature extractor (e.g. clip text encoder). In the last component, we generate the stylized point cloud feature of the specific view, which mixes the content feature from a specific view of the point cloud with the text style feature and the complement global point cloud feature. After all the steps, we project the stylized point cloud features into a 2D stylized view.

\subsection{Point Cloud Construction}
Given a group of images from a specific scene, we can estimate the relative image pose via COLMAP\cite{schoenberger2016sfm}. After this, we can calculate the full-depth map from MVS\cite{schoenberger2016mvs} to construct a 3D point cloud for this scene. Similar to \cite{huang_2021_3d_scene_stylization}, rather than build a point cloud from the image level features, we downsample it into a 3D point cloud of features given the extracted 2D feature map from VGG pre-trained model. 

\subsection{Point Cloud Stylization}
Our methods insert the style into the point cloud descriptor by changing the distribution of content features. Specifically, a Linear transformation module\cite{li2018learning} will predict a transformation matrix $T$ by matching the covariance statistic of content features and text style features. Given the feature vectors of the point cloud and text style embedding, the modulated point cloud will be computed in the equation below.
\begin{gather}
    f^d_p = T(f^c_p - \bar{f^c_p}) + \bar{f^s}
\end{gather}
where $\bar{f^c_p}$ is the mean of feature in the point cloud${f^c_p}$ , $\bar{f^s}$ is the mean of text style features $f^s$, $f^d_p$ is transferred feature of point cloud. 
In the previous work, the reference images provide the multi-scale features extracted from different layers of a pre-trained VGG. The feature of point cloud extracted from multi-view images by a three-layer pre-trained VGG encoder, each point's feature represents the surrounding local receptive field in an image. The multi-scale features from reference images provide a matching representation scale for point cloud stylization. 

\begin{figure*}[h!]
\centering
    \includegraphics[width=0.7\linewidth]{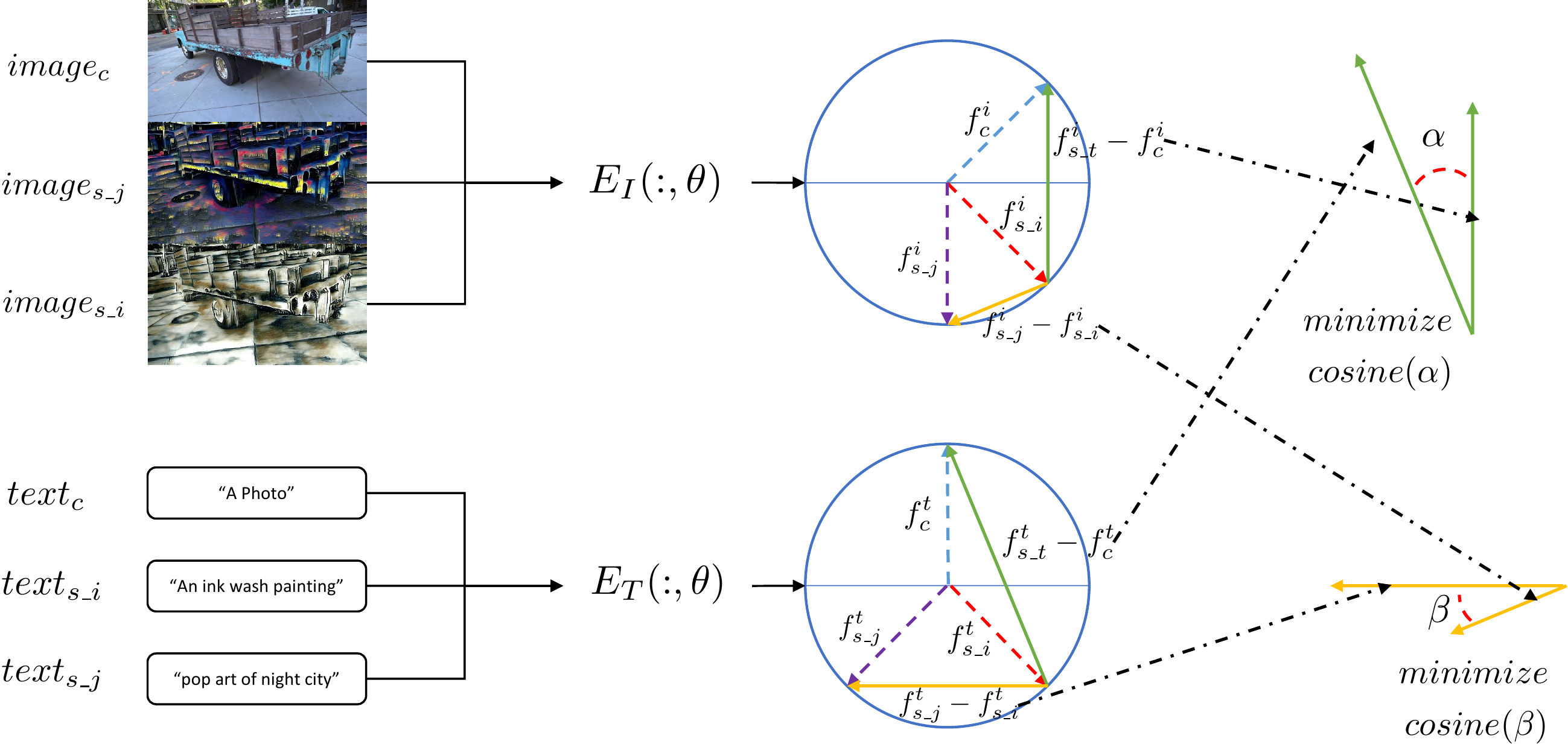}
    \caption{\textbf{Loss function.} To support multiple scenes and styles in one model, only the constraint from source to target will cause the style mix-up. The constraint between styles can strengthen the model to distinguish different styles effectively.}
    \label{loss}
    \vspace{-6mm}
\end{figure*}
However, the connection across the modalities of CLIP is limited to the last layer of the encoders, which only provides the deep feature of text style. Therefore, there is a different representation scale problem between the content and style. To remove this contradiction, we extract a global feature from the point cloud to match the global text style feature by point-wise convolution and max pooling operation. Then, we use the same transformation matrix $T$ to transfer the global feature representation. The global feature will be attached to each point's feature to prepare for view projection. The modulated global feature will be calculated below.
\begin{gather}
    f^c_g = MaxPool(conv(f^c)) \nonumber \\
    f^d_g = T(f^c_g - \bar{f^c}) + \bar{f^s}
\end{gather}
The transformation matrix is calculated from the text style covariance matrix $T^s$ and point cloud content covariance matrix $T^c$. The text features are obtained by feeding the text encoder with different prompt sentences integrating a single text style. Once the style and content features have been calculated, the following convolution layers and a full-connected layer will compute the covariance matrix $T^s$ and $T^c$. Finally, we obtain the transformation matrix $T = T^sT^c$. 
\subsection{Stylized Novel View Synthesis}
After the point cloud has been stylized, the next step is to generate the stylized images with a specified view. View synthesis can be achieved by projecting the point's features to an image plane with a z-buffer, camera pose and intrinsic parameters. Finally, the projected features are mapped to an image by a decoder. 

\textbf{Projection.} Our projector follows the Wiles et al.\cite{wiles2020synsin}, and project the point cloud features into a 2D plane to generate a feature map. In the project process, a z-buffer will accumulate the K-sorted closet points for a pixel. A point will affect $r$ radius pixels on the image plane for better back-propagation and optimization of the decoder.

\textbf{Decoder.} The Decoder maps the projected feature map to an RGB image. The decoder is implemented by the convolution network, following the design of U-net\cite{DBLP:journals/corr/RonnebergerFB15}. It includes the down-sampling and up-sampling operations.

\subsection{Loss Function}
\subsubsection{Style Loss}
\hspace{0.7em} To guide the content scene to follow the semantics of text style, StyleGAN-NADA\cite{gal2021stylegannada} proposed a directional loss that aligns the direction distance of the CLIP feature between the text-image pairs of source and target. The direction loss is given by:
\begin{gather}
    \Delta T = E_T(text_{s}) - E_T(text_{c}), \nonumber \\
    \Delta I = E_I(f(\{P_n^{s}\}^N_{n = 1})) -  E_I(image_{c}), \nonumber \\
    { \mathcal{L}_{dir} = 1 - \dfrac{\Delta I \cdot \Delta T}{|\Delta I||\Delta T|}}
\end{gather}

\noindent where $ \{P_n^{s}\}^N_{n = 1}$ is stylized point cloud, $E_T$ is CLIP text encoder, $E_I$ is CLIP image encoder, $t_{target}$ is the text style, $t_{source}$ is ``a Photo". $f$ is an operation that projects the points to the associate view of the ground truth image and renders the transferred image with a decoder.

To improve the local texture of transferred images, CLIPStyler\cite{kwon2021clipstyler} proposed a patch-based CLIP directional loss. Specifically, the model will randomly crop several patches from rendered $image_{s}$. The size of cropped images is fixed, and following the random perspective, augmentation will be applied on the N cropped patches $\hat{image_{s}^i}$. To alleviate some patch images that are easier to minimize the CLIP loss, the model will reject the patch that the $l_{patch}^i$ value is larger than the specific threshold $\tau$. The PatchCLIP loss is defined as below:
\begin{gather}
    image_{s} = f(\{P_n^{sj}\}^N_{n = 1}), \Delta T = E_T(text_{s}) - E_T(text_{c}), \nonumber \\
    \Delta I = E_I(aug(\hat{image_{s}^i})) -  E_I(image_{c}), \nonumber \\
    l_{patch}^i = 1 - \dfrac{\Delta I \cdot \Delta T}{|\Delta I||\Delta T|}, \mathcal{L}_{patch}^i = \dfrac{1}{N}  \sum_{i}^{N} R(l_{patch}^i,\tau) \nonumber \\
    where R(s,\tau) = \left\{
    \begin{array}{ll}
             0 ,  if  s \leq \tau \\
             s ,  otherwise
    \end{array}\right.
\end{gather}
When we directly adopt the style matching loss of CLIPstyler \cite{kwon2021clipstyler} for multiple text styles transfer, the different styles are easy to mix because the previous PatchCLIP loss only constrains the directional distance from source to target and has no constraint between the different styles. To solve this issue, we propose a directional divergence loss. In a batch, we randomly sample N pairs data and minimize the $\mathcal{L}_{dir}$ loss between the different style data pairs. The following equation describes this loss:
\begin{gather}
    \Delta T = E_T(text_{s\_i}) - E_T(text_{s\_j}), \nonumber\\
    image_{s\_i} = f(\{P_n^{si}\}^N_{n = 1}), image_{s\_j} = f(\{P_n^{sj}\}^N_{n = 1}) \nonumber \\
    \Delta I = E_I(image_{s\_i}) -  E_I(image_{s\_j}), \nonumber\\
    \mathcal{L}_{dir} = 1 - \dfrac{\Delta I \cdot \Delta T}{|\Delta I||\Delta T|},
\end{gather}
where $text_{s\_i}$ and $text_{s\_j}$ are different text styles from the dataset, $\{P_n^{si}\}^N_{n = 1}$ and $\{P_n^{sj}\}^N_{n = 1}$  are point cloud that has been transferred by different style.
Further, we project different stylized views in a batch to assist the model coverage faster and more robustly. If we use the equation above to measure the similarity of the text-image directional distance between different styles, the content disparity of different views will also be included. By calculating the similarity between different views, we successfully remove this noise in the loss function. 
\begin{gather}
    f_c^{i} = E_I(image_{c\_i}), f_c^j = E_I(image_{c\_j})\nonumber\\
    \mathcal{L}_{cd} = 1 - \dfrac{f_c^{i} \cdot  f_c^j}{| f_c^{i}|| f_c^{j}|},
\end{gather}
where $image_{c\_i}$ and $image_{c\_j}$ are content image pairs used for point cloud construction before. Altogether, our style loss function is 
$\mathcal{L}_s =  \mathcal{L}_{patch} + \mathcal{L}_{dir} -  \mathcal{L}_{cd}.$
\subsubsection{Content Loss}

\hspace{0.8em} The preservation of content information is ensured by the VGG perceptual loss $L_{feat}$ between the synthesized image and the ground truth image. An RGB pixel level L1 loss is used to stabilize the model in the training stage. Finally, the content loss is $ \mathcal{L}_c = \lambda_{feat}\mathcal{L}_{feat} + \lambda_{rgb}\mathcal{L}_{rgb}.$
\section{Experiments}
\begin{figure*}
    \centering
    \includegraphics[width=0.85\textwidth]{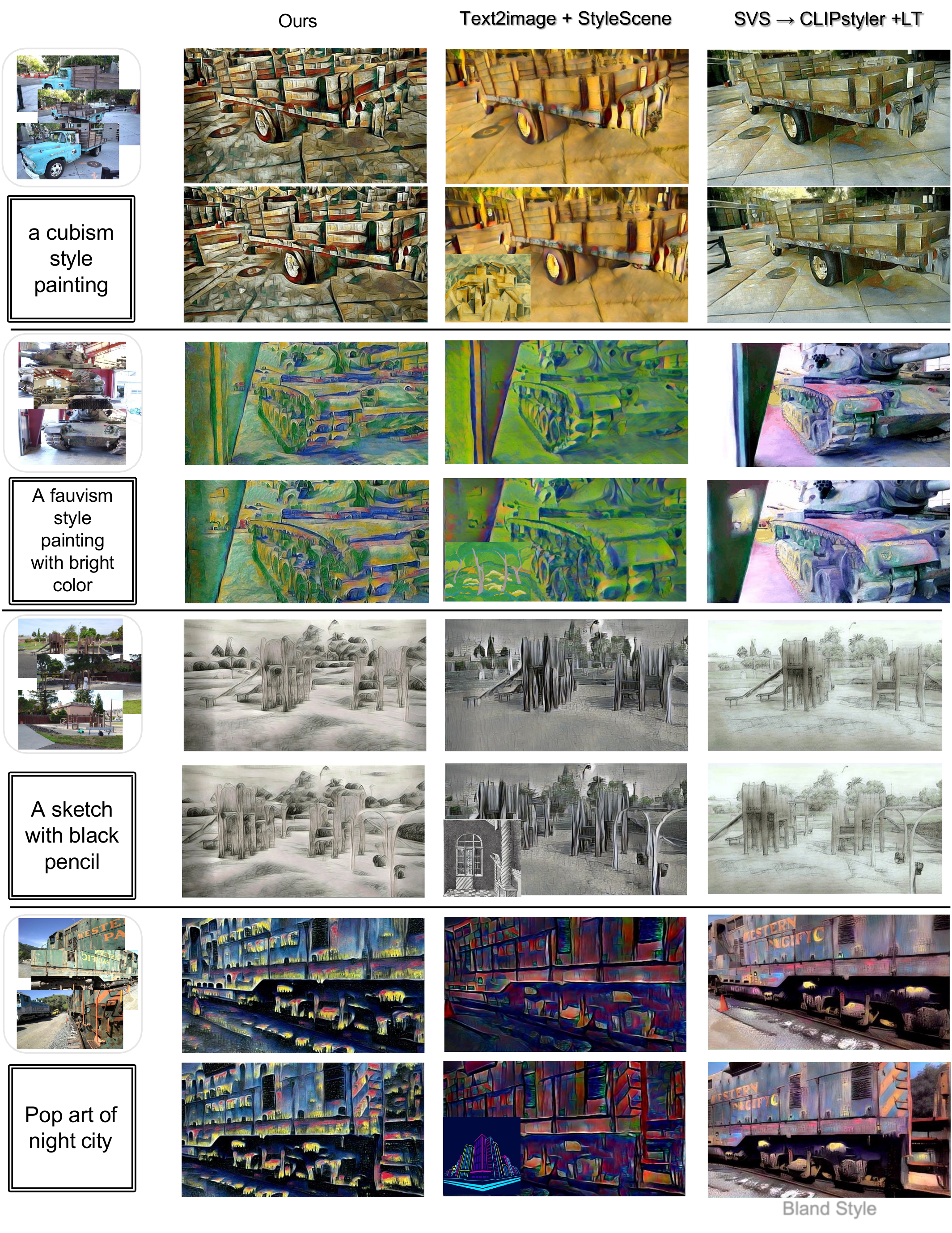}
    \captionof{figure}{\textbf{Comparisons to 2D image text conditional methods and Stylescene.} We compare the images generated from stylizing the novel views with our model.}
    \label{results}
    \vspace{-6mm}
\end{figure*}
To align the experimental settings with the previous method \cite{Riegler2020FVS}, We conduct our experiments on the dataset\cite{10.1145/3072959.3073599}. In this dataset, we split the scenes into training and testing sets. We train our model on sixteen scenes and test the generalization on the four hold-out testing scenes "M60, Truck, Train, and Playground". Besides, our text prompts for stylization are kept consistent among training and test sets.
\subsection{Qualitative results}
To compare the effect of novel stylized views between different methods, we stylized the whole 3D scenes from different methods and sample the stylized views from the same camera poses for comparison.

Due to that StyleScene is conditioned with style images rather than text prompts in our model, we search for the most matched style images to our text prompts for StyleScene to generate the comparable stylized images. And for CLIPstyler\cite{kwon2021clipstyler}, we modify the original AdaIN \cite{huang2017adain} to the more advanced linear style transfer \cite{li2018learning} for a fair comparison, which is kept consistent for StyleScene and our method. Because CLIPstyler\cite{kwon2021clipstyler} only supports a single scene and cannot prevent the issue of style mixing effect, mentioned in Figure~\ref{mixing}, we need to train different models to enumerate all the combinations of different scenes and text styles, in other words, one model for stylizing a single scene with the specific text style. However, our method supports multi-scene, multi-style and generalizable to novel scenes within one model training. Figure~\ref{results} shows the qualitative comparison of our approach with novel view syntheses to the 2D text stylization method and stylescene\cite{huang_2021_3d_scene_stylization}. With the geometry information augmenting our 3D stylization, our model generates more preferable stylization results, and better consistency across the views, which is more stable than the 2D method.

\begin{table}[ht!]
  \vspace{-3mm}
  \begin{center}
    \label{tab:table1}
    \begin{tabular}{l|c|c|c} 
      \textbf{Dataset} &  StyleSene & SVS$\rightarrow$CLIP+LT & \textbf{Ours} \\
      \hline
      Truck       & 0.2371 & 0.2651   & 0.2849\\
      M60         & 0.2362 & 0.2625   & 0.2874\\
      Playground  & 0.2360 & 0.2659   & 0.2822\\
      Train       & 0.2344 & 0.2600   & 0.2859\\
      \hline
      Average     & 0.2359 & 0.2633   & 0.2851 \\
    \end{tabular}
    \caption{\textbf{CLIP Score.} We compare the stylized image of our method with other 2D methods to determine whether our approach better matches the semantic text style.}
    \label{CLIP_score}
  \end{center}
  \vspace{-10mm}
\end{table}

\subsection{Quantitative results}
\begin{figure*}
\centering
    \includegraphics[width=0.8\textwidth]{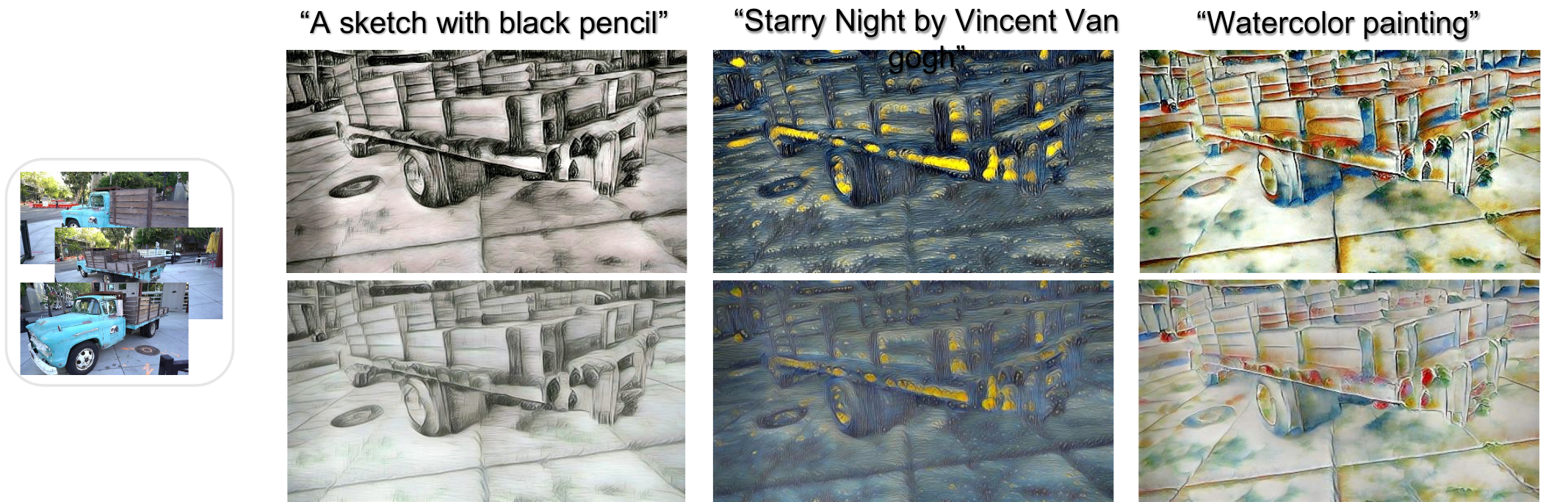}
    \caption{\textbf{Non-global and global feature comparison.} We compare the stylization results between the non-global and global features on the Tank and Temple dataset\cite{10.1145/3072959.3073599}. The top-line images show the results with global information, and the bottom-line photos are not.}
    \label{comparsion}
    \vspace{-6mm}
\end{figure*}
\textbf{User Study.} 
\begin{figure}
    \includegraphics[width=\linewidth]{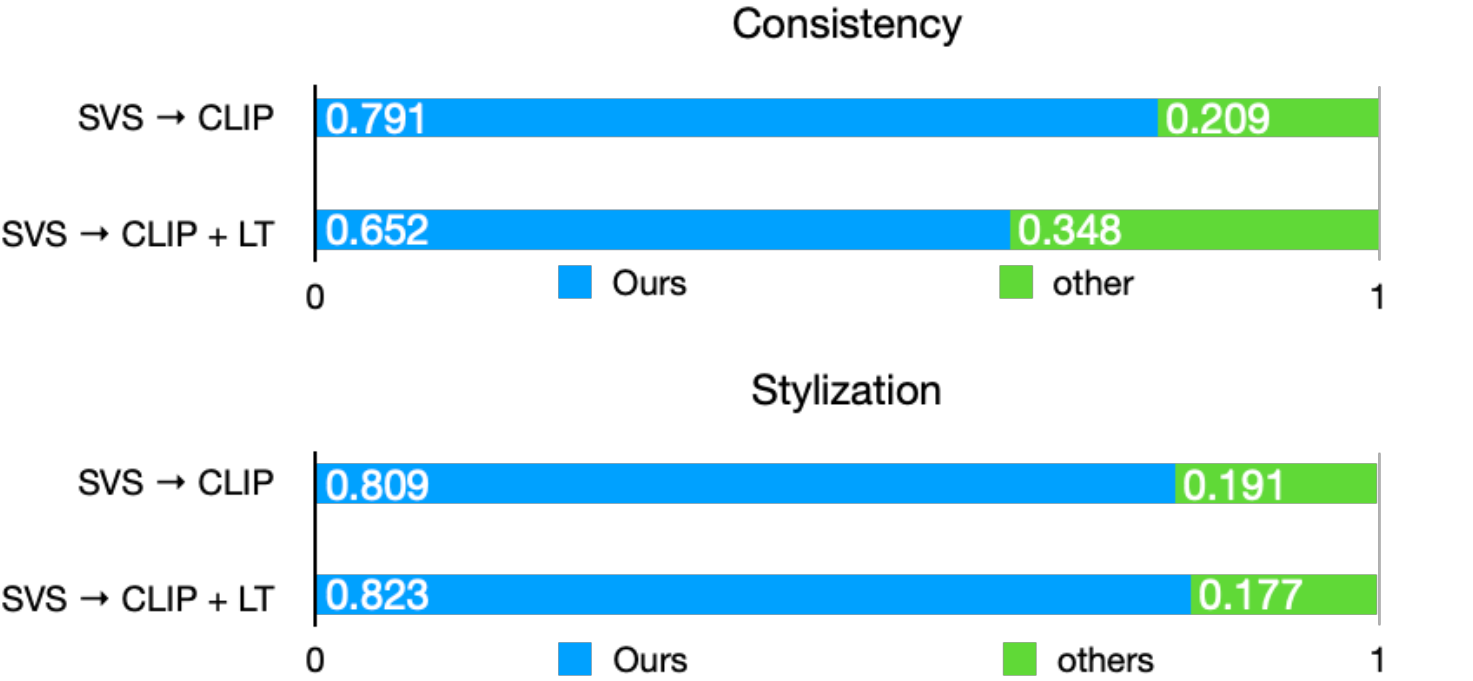}
    \caption{User study. We conduct the user to compare our method with other approaches from consistency and stylization perspectives.}\label{User_study}
    \vspace{-4mm}
\end{figure}
In this qualitative experiment, we create an anonymous voting pool for comparing different methods. We generate 21 different stylized scenes and convert them to GIF format via the order of camera poses. In the voting section, the users are asked to choose the preferable stylized scenes w.r.t. the more faithful style effect or better view consistency. In total, we have 60 participants successfully finishing the voting questionnaire.
\begin{table}[ht!]
  \begin{center}
    \begin{tabular}{l|c|c|c} 
      \textbf{Dataset} &  StyleScene & SVS$\rightarrow$CLIP+LT & \textbf{Ours} \\
      \hline
      Truck       & 0.0835 & 0.0978   & 0.0827\\
      M60         & 0.0939 & 0.1037   & 0.0644\\
      Playground  & 0.0762 & 0.0933   & 0.0441\\
      Train       & 0.0827 & 0.1130   & 0.0818\\
      \hline
      Average     & 0.0841 & 0.1019   & 0.0683 \\
    \end{tabular}
    \caption{\textbf{Short-range consistency.} We use the (t-1)-th and t frame of video to measure the color variance by RMSE. }
    \label{short-range consistency}
    \end{center}
  \vspace{-4mm}
\end{table}
\begin{table}[ht!]
  \begin{center}
    \begin{tabular}{l|c|c|c} 
      \textbf{Dataset} &  StyleScene & SVS$\rightarrow$CLIP+LT & \textbf{Ours} \\
      \hline
      Truck       & 0.1065 & 0.1109   & 0.1007\\
      M60         & 0.1150 & 0.1239   & 0.0754\\
      Playground  & 0.1152 & 0.1091   & 0.0631\\
      Train       & 0.1095 & 0.1285   & 0.1066\\
      \hline
      Average     & 0.1116 & 0.1181   & 0.0864 \\
    \end{tabular}
    \caption{\textbf{Long-range consistency.} We use the (t-7)-th and t frame of the video to measure the color variance by RMSE. }
    \label{long-range consistency}
  \end{center}
  \vspace{-8mm}
\end{table}
As shown in Figure~\ref{User_study}, The users deem that our approach reaches better consistency and conforms more to the target style.
\textbf{Stylization Quality.}
To quantify the stylization quality, we calculated the cosine similarity between the CLIP embedding of output stylized images and associated targets style text defined by CLIPstyler\cite{kwon2021clipstyler}. To better measure the local texture style transformation quality, we randomly crop 64 patches before calculating the image CLIP embedding. 
\begin{table}[ht!]
  \begin{center}
    \begin{tabular}{l|c|c|c} 
      \textbf{Dataset} &  global & non-global \\
      \hline
      Truck       & 0.2849 & 0.2808   \\
      M60         & 0.2874 & 0.2751   \\
      Playground  & 0.2822 & 0.2678   \\
      Train       & 0.2859 & 0.2810   \\
      \hline
      Average     & 0.2851 & 0.2761    \\
    \end{tabular}
    \caption{\textbf{CLIP score of global and non-global.} By enhanced the global feature transformation, the stylization views achieve higher CLIP score, and better match text described style.}
    \label{clip score of global and non-global}
  \end{center}
  \vspace{-11mm}
\end{table}

The comparison with other 2D methods is shown in Table~\ref{CLIP_score}.
With the advantage of 3D geometry, we achieve better stylization results with higher CLIP scores than the 2D method and maintain the content of a scene.
\textbf{View Consistency.} To measure inconsistency between the pair of stylized views, we reproject the pixel to the 3D space and project back to another plane by the camera intrinsic and extrinsic of a pair of views. By doing this, we can measure the color changing of a pair of pixels in different views projected from the same point and compute RMSE to measure these color differences as a consistency metric. Similar to Huang \etal \cite{huang_2021_3d_scene_stylization}, we calculate the RMSE of the adjacent frames of a video to quantify the short-range consistency and the viewpoints of (t - 7)-th and t-th video frames to quantify the long-range consistency. Finally, we average the results from 21 different styles for all test scenes and report the mean value in Table~\ref{short-range consistency} and Table~\ref{long-range consistency}. For either the short-range or long-range consistency, the proposed method reaches the lower RMSE values, which means the color of pixels projected onto different views from a point variant is smaller than the 2D method.
\subsection{Ablation studies}
\textbf{Compare with CLIPNeRF.} To compare with the alternative solution of 3D scene stylization, we conduct a straightforward comparison between our method and the recent CLIPNeRF \cite{wang2021clip} on the same scene and the same text style prompts. It is obvious CLIPNeRF seems to only learn a color shift for the given style, which is not feasible compared with ours.

\textbf{Effect of global feature transformation.} We compare the stylization results with the non-global and global operations and calculate the CLIP score to measure whether the global information helps the model better match the text-described style. In Figure~\ref{comparsion}, with the global information, we obtained a much higher contrast and full texture details on the ground and the object surface. Without the global transferred feature, point-wise feature style transformation causes the obscure or indistinct of the truck and tank in the figure. We also measure the CLIP score of the render results with global and non-global style transformation in Table~\ref{clip score of global and non-global}.

%
\section{Conclusions}
In this paper, we introduce the text conditional stylization on 3D scenes generating novel views from a set of images with a semantic text style. The model effectively distinguishes the different text styles by involving new directional distance constraints. Integrating the global transferred feature to the projected feature map enhances the model performance on fine details. We demonstrated the efficacy of our methods through extensive qualitative and quantitative studies.

\clearpage
{\Large \noindent \textbf{Supplementary Material}}
\section{Implementation details}
\textbf{Point Cloud Global Feature Extractor.} A lightweight module extracts the global feature. It implemented as several 1 x 1 convs to increase the channel of point cloud's features from 256 to 1024. The following MaxPooling operation aggregates the all features from the point cloud to generate a unified global representation. To use the same Transformation Matrix $T$ calculated from the point cloud feature and style embedding, we have to compress the global feature to 256 channels by 1 x 1 convs.

\textbf{Neural Decoder.} The architecture of the decoder follows the design of U-Net\cite{DBLP:journals/corr/RonnebergerFB15}; it first receives 256 channels projected feature map. Then encoder part of U-Net downsamples the feature map by a series of convolution and AvgPool layers. It forces the model not only to memorize the content of a scene. The subsequently transposed convolution layers up-sample the feature maps to the three-channel images. All layers in the U-Net have a kernel size of 3 x 3 with ReLU non-linear function except the skip convolution.

\textbf{Training.} We first train the decoder without inserting the style into the point cloud feature with a batch size of $4$ and a learning rate of $0.0001$. The $\lambda_{rgb}$ pixel of reconstruction loss $\mathcal{L}_{rgb}$ is 0.005, and the $\lambda_{feat}$ of feature perception loss $\mathcal{L}_{feat}$ is 1.0. We then train the style transformation module with a batch size of 4 and a learning rate of 0.0001. To coverage faster of the model, we select different views from a scene in a batch. We also involve the $\mathcal{L}_{tv}$ and $\mathcal{L}_{gs}$ loss from \cite{kwon2021clipstyler} to alleviate the side artifacts and maintain the global content of an image. For hyperparameters, we set the $\lambda_{rgb}, \ \lambda_{feat}, \ \lambda_s$ and $\ \lambda_{tv}$ as $5 \times 10^{-3}, 1, 1.5 \times 10 , 1.3 \times 10^{-6}$ respectively.  We use the Adm optimizer with $\beta_1 = 0.9$ and $\beta_2 = 0.9999$ for all network training.

\textbf{CLIP Style Implementation Details.} The input of the CLIP model is an image with 224 $\times$ 224 resolution. We have to resize the image before feeding it into the CLIP's image encoder. Following the \cite{kwon2021clipstyler}, we randomly cropped 64 patches with the size of 96 and applied random perspective augmentation on each patch. For the threshold rejection $\tau$, we set $\tau$ as 0.7. To measure the directional divergence of different styles, we randomly sample 80\% of all pairs of different styles to reduce the computation cost. 

\textbf{Style Transformation Module.} Following the Linear Transformation\cite{li2018learning} module, we compress the feature of the point cloud from 256 to 64 by an MLP layer. By reducing the feature's dimension, we accelerate the covariance matrix computation of the point cloud and transformation process.
For the text style, we insert the text prompt into 79 template sentences to get the different perspectives representation of the style description. We then compress the feature size from 512 to 64 to calculate the covariance matrix.

\begin{center}
    \centering
    \captionsetup{type=figure}
    \includegraphics[width=0.6\linewidth]{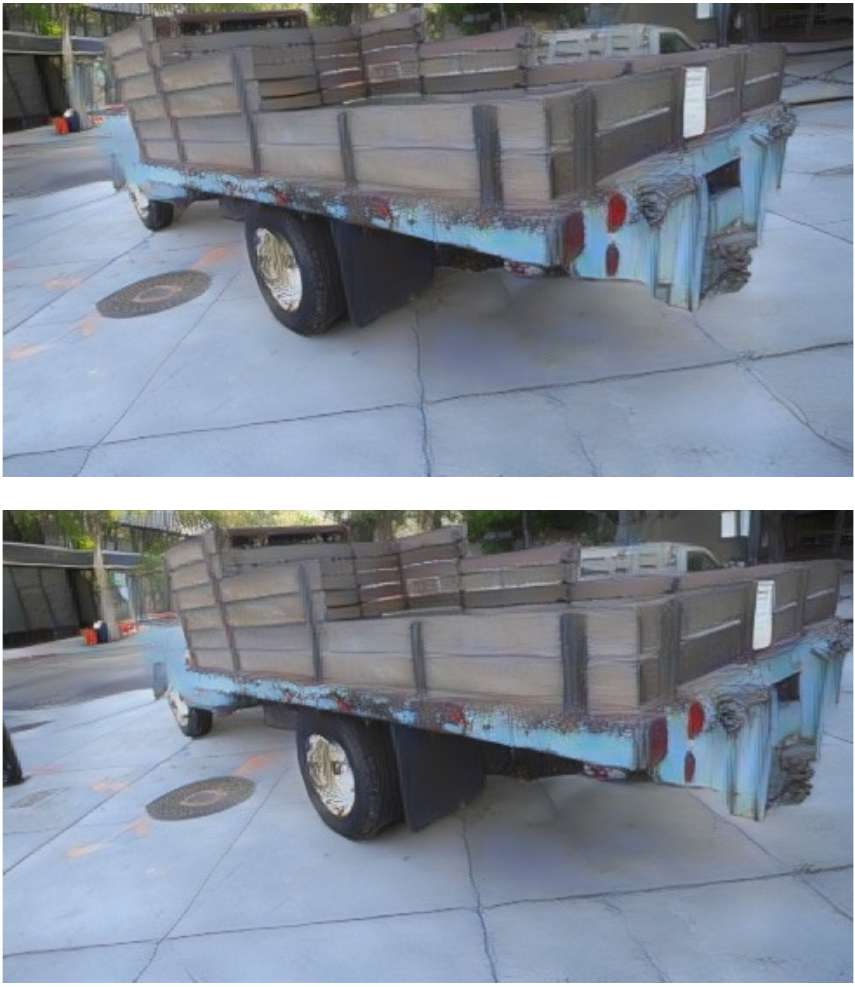}
    \captionof{figure}{Synthesized views without style}
    \label{without_style}
\end{center}%
After the style of the point cloud has been translated, we uncompress the feature from 64 to 256 channels. The global representation of the point cloud is also translated by the same transformation matrix $T$. The translated global feature will be attached to each point cloud feature, and the combined feature will be compressed to 256 dimensions.

\textbf{2D Method Experiment.} Since we do not have the ground truth image of the synthesized views, we have to generate the associate view without the style first, as shown in Figure~\ref{without_style}, and use the 2D method to stylize all of the views. In the training stage of the 2D method, each model only supports a single style and image. Therefore, we train the model with the multi-images from a single scene and a style.

\section{Limitation and Future Direction}
First, our model relies on the structure from motion to generate the point cloud of a scene, and it requires lots of images of overlapped views. If the pictures we provide have no overlap view with each other or the number of photos is minimal, the sfm algorithm cannot generate a good point cloud to represent a scene. Second, we extract the feature from the 2D images by a pre-trained VGG encoder as a feature of the point cloud; the geometry information is not inserted into the feature, which causes a consistency problem across the views. Third, we train the model with a batch size of 4, which means we have to cache four different stylized point clouds, and it consumes a lot of memory. With better memory consumption optimization, we can enlarge the batch size to make the model coverage faster. For example, we can pre-calculate the projected feature and record the point cloud's index. We then only need to cache the projected stylized feature, and we do not need to copy other features to reduce the memory consumption of the GPU.

\section{Additional Results}
\twocolumn[{%
\renewcommand\twocolumn[1][]{#1}%
\begin{center}
    \centering
    \captionsetup{type=figure}
    \includegraphics[width=0.9\linewidth]{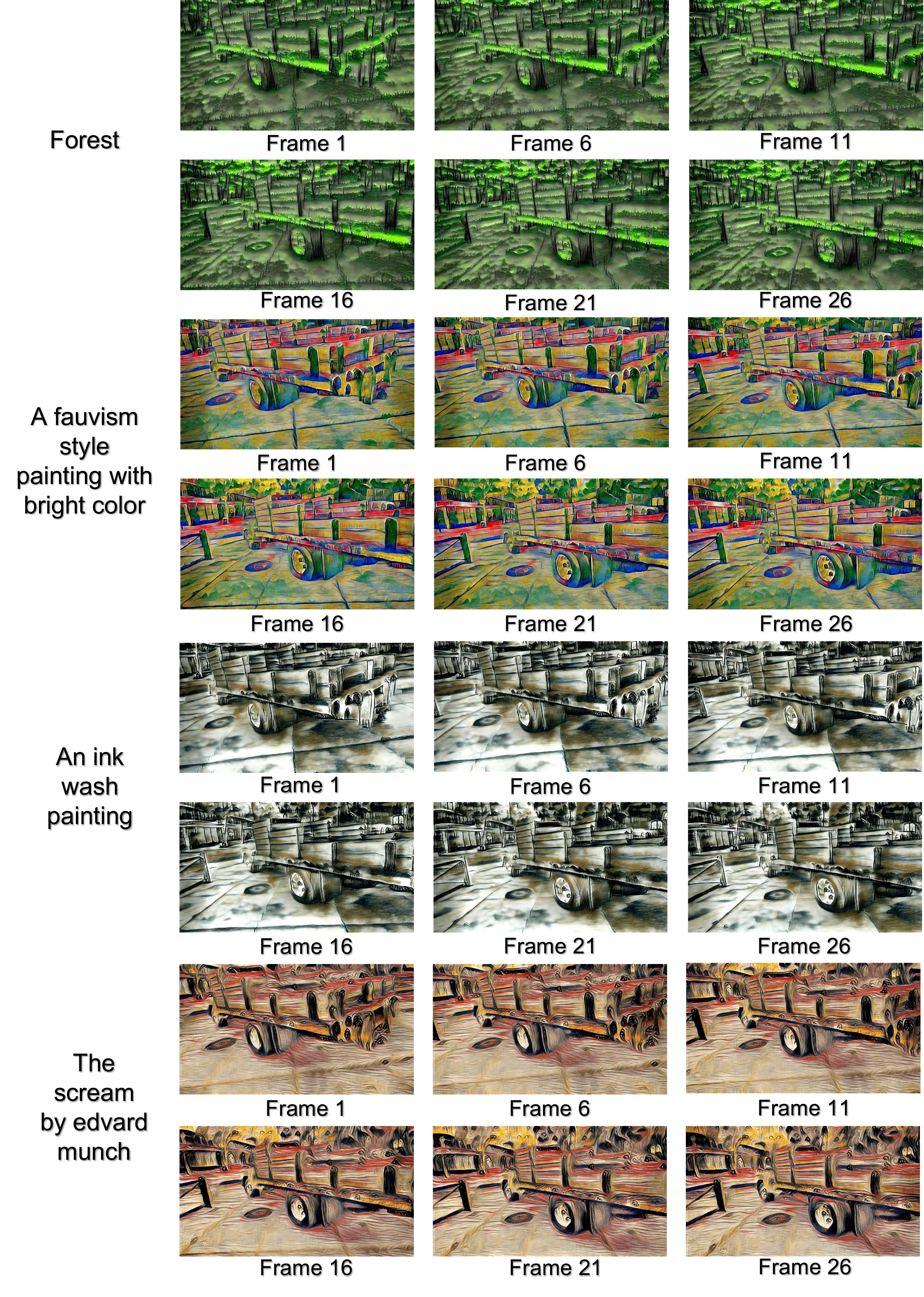}
    \captionof{figure}{additional results}
    \label{supply1}
\end{center}%
}]

\twocolumn[{%
\renewcommand\twocolumn[1][]{#1}%
\begin{center}
    \centering
    \captionsetup{type=figure}
    \includegraphics[width=0.88\linewidth]{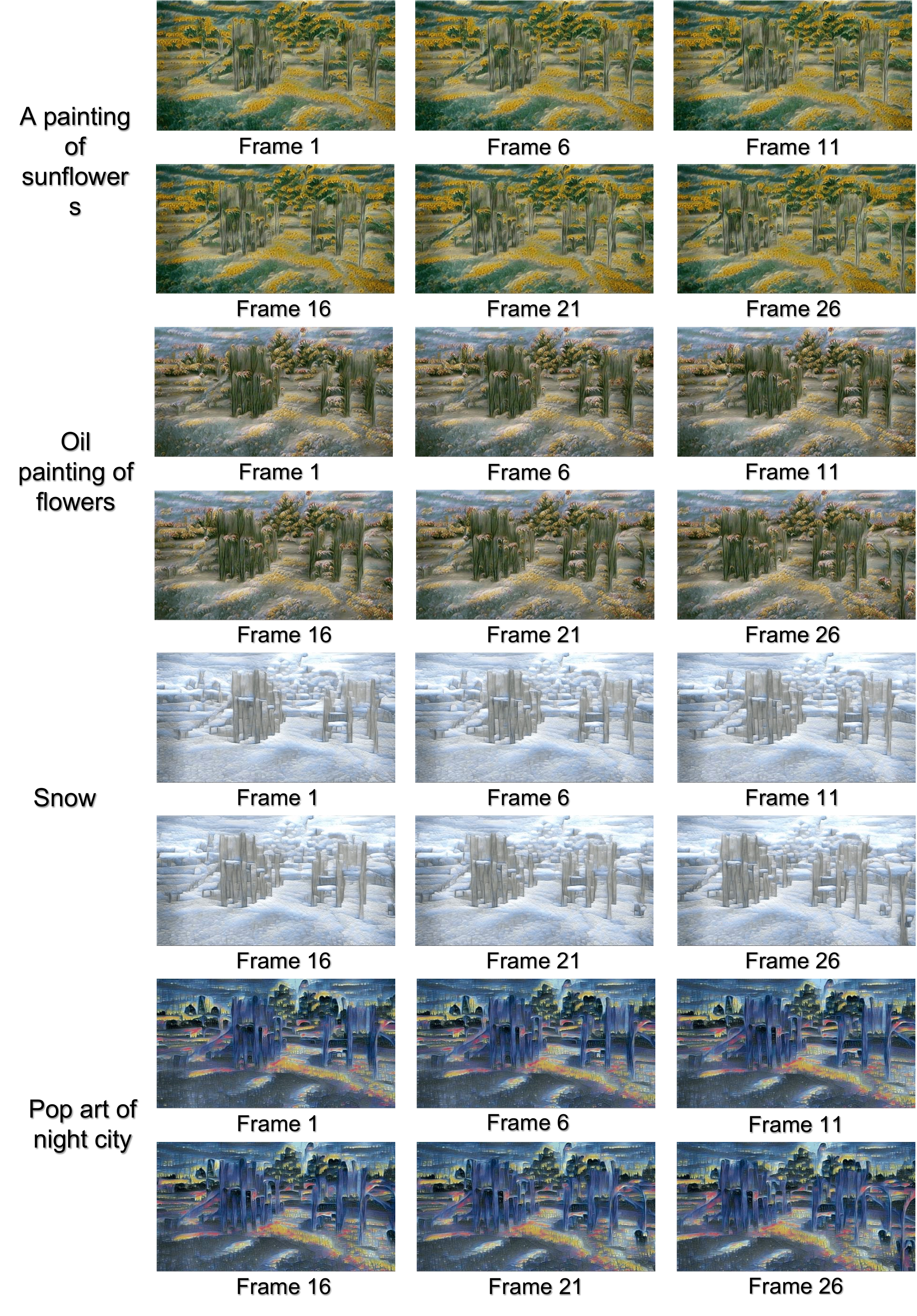}
    \captionof{figure}{additional results}
    \label{supply2}
\end{center}%
}]

\twocolumn[{%
\renewcommand\twocolumn[1][]{#1}%
\begin{center}
    \centering
    \captionsetup{type=figure}
    \includegraphics[width=0.88\linewidth]{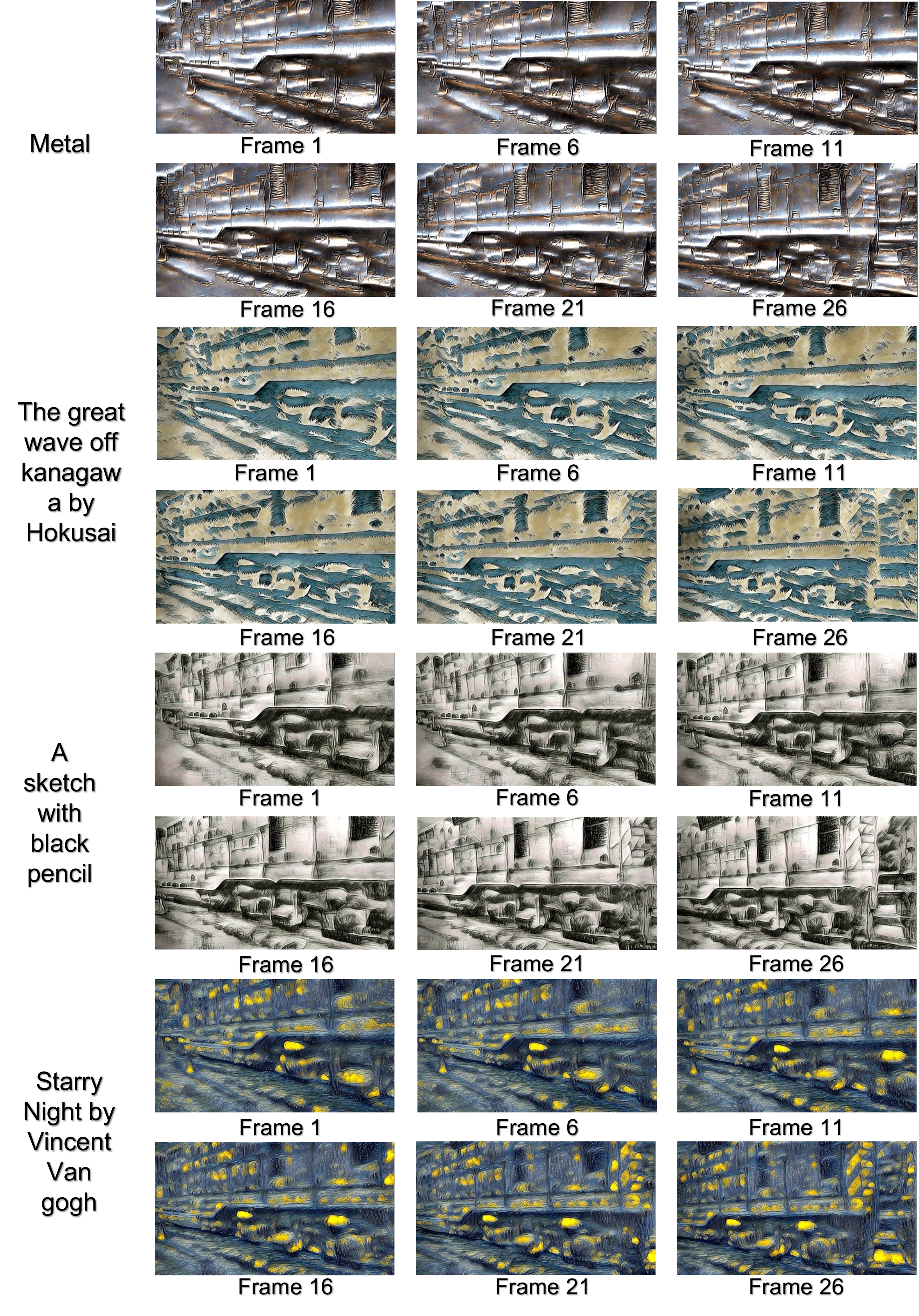}
    \captionof{figure}{additional results}
    \label{supply3}
    
\end{center}%
}]

\twocolumn[{%
\renewcommand\twocolumn[1][]{#1}%
\begin{center}
    \centering
    \captionsetup{type=figure}
    \includegraphics[width=0.92\linewidth]{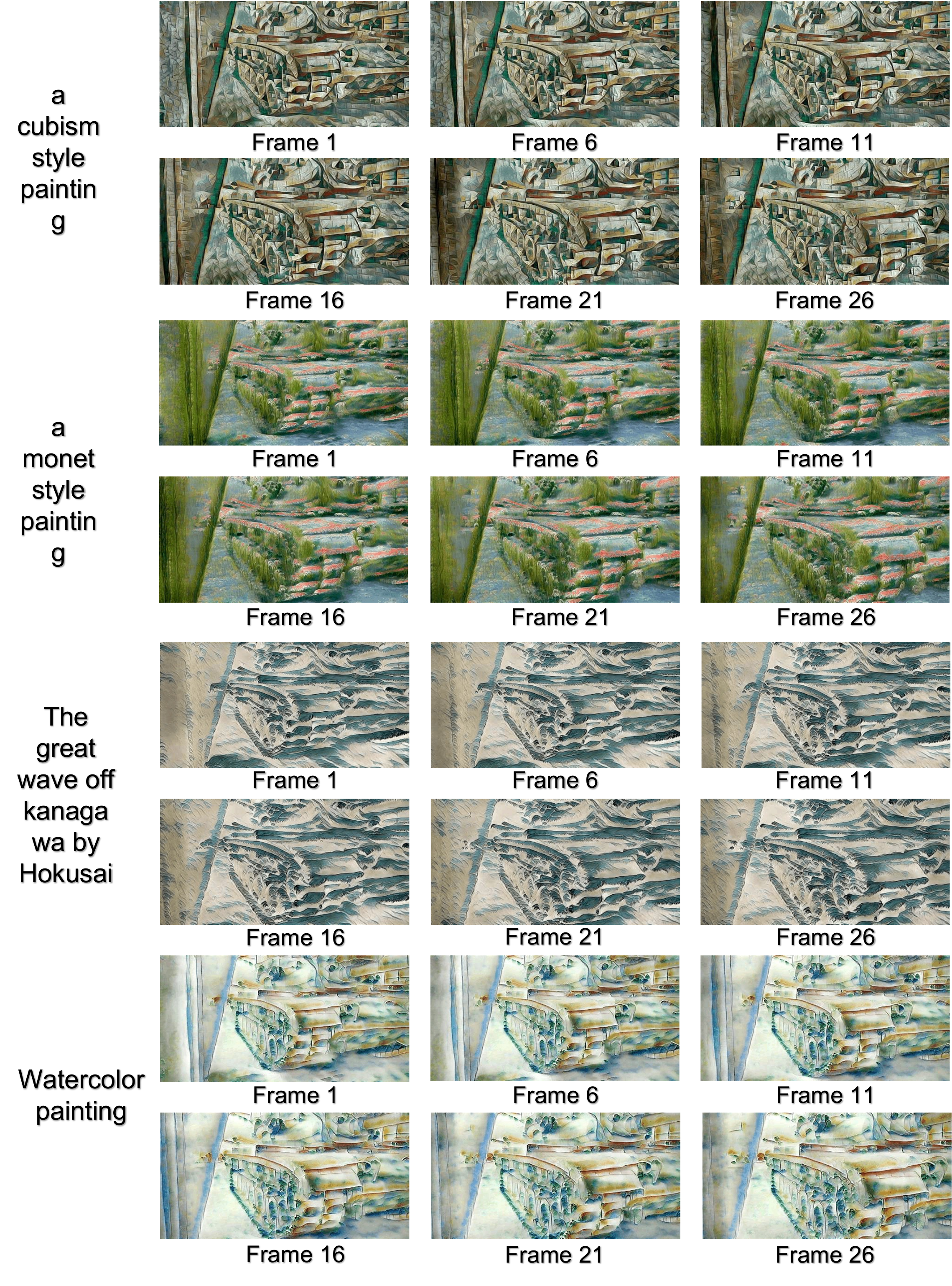}
    \captionof{figure}{additional results}
    \label{supply4}
    
\end{center}%
}]

{\small
\bibliographystyle{ieee_fullname}
\bibliography{egbib}
}

\end{document}